\pdfoutput=1
\documentclass[conference]{IEEEtran}
\IEEEoverridecommandlockouts
\usepackage{cite}
\usepackage{amsmath,amssymb,amsfonts}
\usepackage{algorithmic}
\usepackage{graphicx}
\usepackage{textcomp}
\usepackage{xcolor}
\def\BibTeX{{\rm B\kern-.05em{\sc i\kern-.025em b}\kern-.08em
    T\kern-.1667em\lower.7ex\hbox{E}\kern-.125emX}}
    
\makeatletter
\newcommand{\linebreakand}{
  \end{@IEEEauthorhalign}
  \hfill\mbox{}\par
  \mbox{}\hfill\begin{@IEEEauthorhalign}
}
\makeatother

\begin{document}

\title{Application for White Spot Syndrome Virus (WSSV) Monitoring using Edge Machine Learning \\
}
\author{\IEEEauthorblockN{Lorenzo S. Querol}
\IEEEauthorblockA{\textit{Computer Technology Department} \\
\textit{De La Salle University}\\
Manila, Philippines \\
renzo\_querol@dlsu.edu.ph}
\and 
\IEEEauthorblockN{Macario O. Cordel, II}
\IEEEauthorblockA{\textit{Computer Technology Department} \\
\textit{De La Salle University}\\
Manila, Philippines \\
macario.cordel@dlsu.edu.ph}
\linebreakand
\IEEEauthorblockN{Dan Jeric A. Rustia}
\IEEEauthorblockA{\textit{Greenhouse Horticulture and Flower Bulbs Business Unit} \\
\textit{Wageningen University \& Research}\\
Wageningen, The Netherlands \\
dan.rustia@wur.nl}
\and
\IEEEauthorblockN{Mary Nia M. Santos}
\IEEEauthorblockA{\textit{Aquaculture Research and Development Division} \\
\textit{National Fisheries Research and Development Institute}\\
Quezon City, Philippines \\
nia.santos@nfrdi.da.gov.ph}
}

\maketitle

\begin{abstract}
The aquaculture industry, strongly reliant on shrimp exports, faces challenges due to viral infections like the White Spot Syndrome Virus (WSSV) that severely impact output yields. In this context, computer vision can play a significant role in identifying features not immediately evident to skilled or untrained eyes, potentially reducing the time required to report WSSV infections. In this study, the challenge of limited data for WSSV recognition was addressed. A mobile application dedicated to data collection and monitoring was developed to facilitate the creation of an image dataset to train a WSSV recognition model and improve country-wide disease surveillance. The study also includes a thorough analysis of WSSV recognition to address the challenge of imbalanced learning and on-device inference. The models explored, MobileNetV3-Small and EfficientNetV2-B0, gained an F$_1$-Score of 0.72 and 0.99 respectively. The saliency heatmaps of both models were also observed to uncover the ``black-box" nature of these models and to gain insight as to what features in the images are most important in making a prediction. These results highlight the effectiveness and limitations of using models designed for resource-constrained devices and balancing their performance in accurately recognizing WSSV, providing valuable information and direction in the use of computer vision in this domain.
\end{abstract}


\section{Introduction}
Shrimp is a priority export commodity in the aquaculture sector. According to the 2021 Fisheries Profile of a government office \cite{BFAR2021}, shrimp production was estimated to be 68,987 metric tons, ranking second and contributing to 22.03\% of the country’s total aquaculture production with a value of US\$519 billion. Although exports have increased significantly, the presence of viral diseases such as White Spot Syndrome Virus (WSSV), Hepatopancreatic Parvovirus, Acute Hepatopancreatic Necrosis Disease, and Yellow Head Virus has affected the production rate, resulting in a loss as high as 40,800 metric tons in exports \cite{Macusi2022}. WSSV is listed by the World Organization for Animal Health as a major viral disease that affects the global shrimp industry. From 2009 to 2018, the average annual losses in the Asian shrimp industry was US\$4 billion \cite{shinn2018asian}. WSSV mainly affects penaeid shrimps\footnote{Penaeidae is a crustacean family well-known for its economic value in aquaculture.} and is caused by WSSV, a virus that is part of the genus Whispovirus and thus part of the Nimaviridae family \cite{Wang2019}. Once infected, the virus causes a rapid outbreak, with dead shrimp emerging near or at the bottom of the pond in 3 to 10 days. The most obvious indication of WSSV infection is calcium deposits, which appear as white spots on the shrimp carapace. Two days after the virus infects the infected shrimp, white patches may start to appear. 

\begin{figure}[!t]
    \centering
    \includegraphics[width=3.45in]{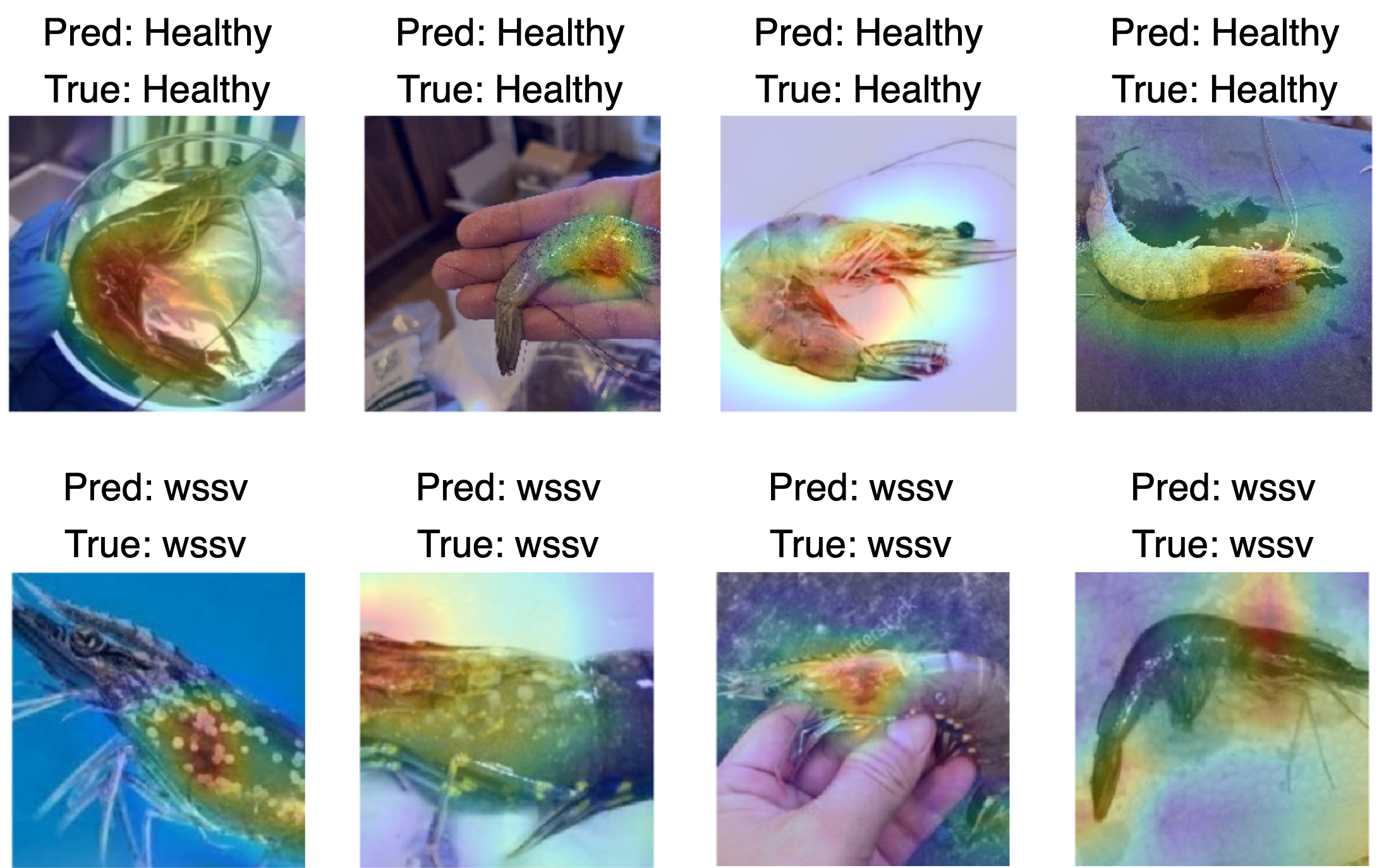}
    \caption{Saliency heatmaps generated by EfficientNetV2-B0. The top row displays image samples of healthy shrimps and the bottom row shows WSSV-infected shrimps. The red areas indicate where the model pays more attention, while the blue areas indicate little attention. In this case, the model pays more attention to areas where WSSV occurs, indicating that the model is learning meaningful patterns from the shrimp images during training.}
    \label{fig:efficientnet-saliency}
\end{figure}

The absence of a recognized cure for the viral infection that causes WSSV makes early detection and prevention crucial to stop its spread. Unfortunately, when a shrimp farm experiences a significant number of deaths in its ponds, the most extreme and unsustainable measure to eliminate the virus is exterminating the remaining shrimp population. To prevent disease transmission, strict biosecurity measures are implemented. However, due to limited on-site equipment, especially in small-scale farms, farmers must take shrimp samples with suspected WSSV infection to a laboratory for testing by professionals. Some farmers may be unaware of the symptoms that indicate a WSSV infection, relying solely on traditional knowledge and experience. This delay in effective disease management or suspicion notification can lead to disease outbreaks near the farm.

Computer vision systems have gained popularity for identifying agricultural diseases \cite{ahmed2019rice, mohanty2016using, sujatha2021performance}, enabling faster diagnosis by detecting indiscernible features. Although farmers can mitigate viral spread with their expertise, this approach might be inefficient and costly due to preparations. In contrast to Polymerase Chain Reaction (PCR) tests used in shrimp disease diagnosis, computer vision remains relatively unexplored \cite{Duong-Trung2020} for preliminary disease screening. Leveraging computer vision can potentially identify shrimp diseases, optimize farm management with standard protocols, and reduce unnecessary costs.

This study addresses the absence of publicly available image datasets in the domain of shrimp disease detection by developing a mobile application for the collection of image samples from shrimp farms. Using the images collected, the training of a recognition model was then conducted to demonstrate its potential in WSSV recognition. This study also recognizes the limited model evaluation in the current literature by utilizing more appropriate evaluation metrics and techniques, offering a deeper understanding of the performance and potential of a computer vision system in shrimp disease detection.

The following are the contributions made by this study:
\begin{itemize}
\item This study presented a mobile application developed to assist in the collection of data from shrimp farms, allowing end users to submit reports containing images flagged by the WSSV recognition model, along with other relevant details such as water, environmental, and geospatial parameters. The mobile application facilitated the collection of valuable data on the occurrence of WSSV throughout the year, allowing future research and analysis in the field of shrimp disease management.

\item This study also presented a computer vision-based solution for WSSV recognition, investigating the use of transfer learning techniques to address limited data in training the WSSV recognition model. Using various evaluation metrics, this study demonstrates the performance and effectiveness of the proposed approach.
\end{itemize}

\section{Related Works}
\subsection{Mobile Application-Based Animal Health Monitoring}
The use of smartphones in low-resource environments is beneficial in increasing the number of reports from farmers and surveillance officers monitoring the spread of livestock diseases \cite{Thumbi2019, George2021, Njenga2021}, as well as the surveillance of transmissible livestock to human diseases \cite{Karimuribo2017}. This strategy in disease surveillance can profoundly impact the animal health sector, paving the way for timely interventions. However, the utilization of mobile application-based monitoring systems in various domains is still an emerging field and its widespread adoption in all fields has not yet been fully established. To our knowledge, in the context of aquaculture diseases, and particularly in the case of monitoring WSSV, there is currently no viable mobile application specifically designed for this purpose. This study aims to develop a mobile application that uses computer vision-based techniques for the detection of WSSV in shrimps to facilitate early detection and reporting.

\begin{figure*}[!t]
    \centering
    \includegraphics[width=0.88\textwidth]{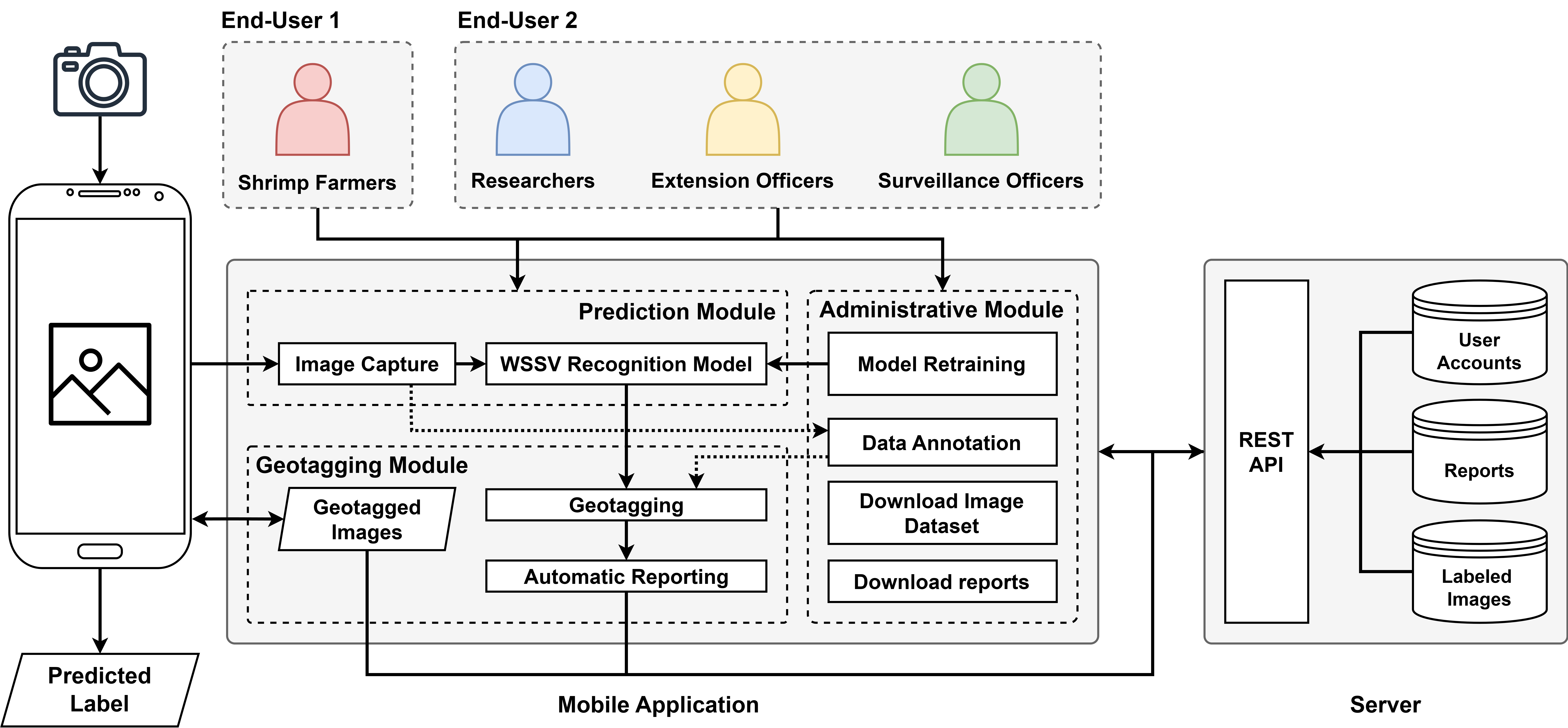}
    \caption{System overview of the mobile application's interaction with the server.}
    \label{fig:system-overview}
\end{figure*}

\subsection{Shrimp Disease Recognition Using Machine Learning}
The first study on shrimp disease detection using computer vision-based methods was conducted using digital signal processing techniques. Sankar and his colleagues \cite{sankar2013white} utilized techniques such as hyperspectral imaging and wavelet transforms, and $k$-Means Clustering for pattern recognition. Although this approach highlights the details of an image, the equipment needed to employ this method can be costly and several factors must be controlled, such as lighting and background, to minimize noise. To improve this approach, subsequent efforts \cite{sucharitha2013identification, anitha2014classifying} used more commonly known image processing techniques. The introduction of classical machine learning algorithms in this domain was carried out by \cite{sucharita2015penaeid} to detect shrimp species using SVM. This led to an interest in the recognition of various shrimp diseases \cite{sankar2017diagnosis} using Local Binary Patterns as the features and $k$-Nearest Neighbors as the classifier. The emergence of deep learning allowed further studies in this domain. The use of techniques such as Convolutional Neural Networks allowed for samples-driven methods to extract more relevant features of shrimp diseases in images. However, only a few works \cite{Duong-Trung2020, Evci2019} have been conducted in this domain, particularly using transfer learning which is known to be effective in dealing with limited datasets.

Recent works \cite{Duong-Trung2020, Evci2019} on deep learning in shrimp disease recognition provide limited evaluation, particularly on the systems' classification performance on imbalanced data samples. This study aims to create a more thorough analysis of the use of a computer vision model for WSSV identification.

\section{Mobile Application Development}
The system illustrated in Fig. \ref{fig:system-overview} comprises two components: the mobile application and the server. The mobile app interacts primarily with the server to save and retrieve images and reports from shrimp farms, reducing the need for constant internet connectivity. The mobile app has three modules: prediction, geotagging, and administrative modules. The prediction module preprocesses input images and runs inference with the WSSV recognition model. The geotagging module determines the user's location, either through GPS or manual input, and reports if the image is predicted to be WSSV-infected. The administrative module handles retraining of the model, image dataset management, reports, and image annotation capabilities. Developed for Android devices, the mobile app uses \textit{TypeScript}, \textit{React Native}, and \textit{Expo}.

The WSSV recognition model is deployed on smartphones using \textit{Open Neural Network Exchange} (ONNX). The app also includes a reporting system allowing users to submit reports containing flagged images recognized by the trained model, along with additional details such as water and environmental parameters. Encouraging users to contribute data through the app creates a valuable dataset for understanding WSSV prevalence and fluctuations over time.

Since WSSV-infected shrimp may show no visible symptoms, relying on white spots as an indicator is unreliable due to similar spots caused by other factors. Despite these limitations, the mobile app's role in disease monitoring and surveillance is crucial, especially for early detection. Gathering data from shrimp farms and integrating production metrics, water quality parameters, and environmental data provides a comprehensive overview of shrimp farming practices and their relationship with disease occurrences.

\section{Computer Vision-Based Detection of WSSV}
\begin{figure}[!t]
    \centering
    \includegraphics[width=3.1in]{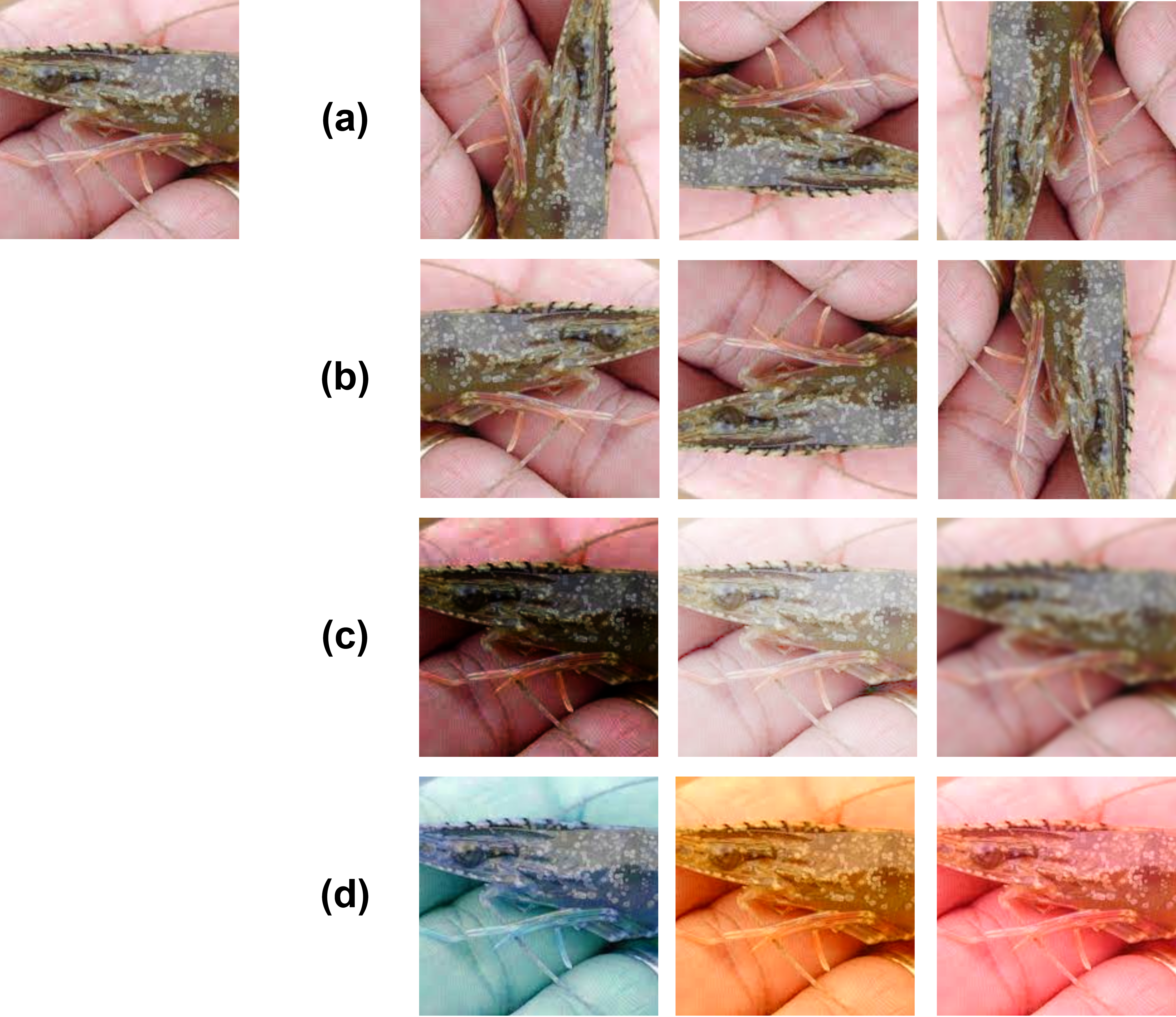}
    \caption{Examples of data augmentation techniques; (a) Rotation; (b) Flipping; (c) Brightness and Blur; (d) Color space transformations might be dangerous when used for augmentation.}
    \label{fig:data-augmentation}
\end{figure}

A dataset comprised of a combination of web-scraped images and smartphone images, taken using different smartphone models, was collated. In total, 447 images were collected, with 411 labeled as healthy and 38 labeled as WSSV-infected. For training and evaluation, the image dataset was split into 80\% for cross-validation and 20\% for the final evaluation of the models. As the dataset size is limited, data augmentation was used and applied to the training samples to increase the number of image samples and create a more diverse dataset for training the recognition model. These augmentation techniques involve transformations that simulate images captured by smartphones, some of which are shown in Fig. \ref{fig:data-augmentation}. Color transformations may potentially affect relevant features of WSSV-infected shrimps. Thus, the transformations applied to the images were ensured not to affect these features. The images were then cropped to a square size region of interest and then resized to a resolution of 224$\times $224 pixels to meet the model input shape requirements. An evaluation was also performed on the test samples.

\begin{figure}[!t]
    \centering
    \includegraphics[width=3.45in]{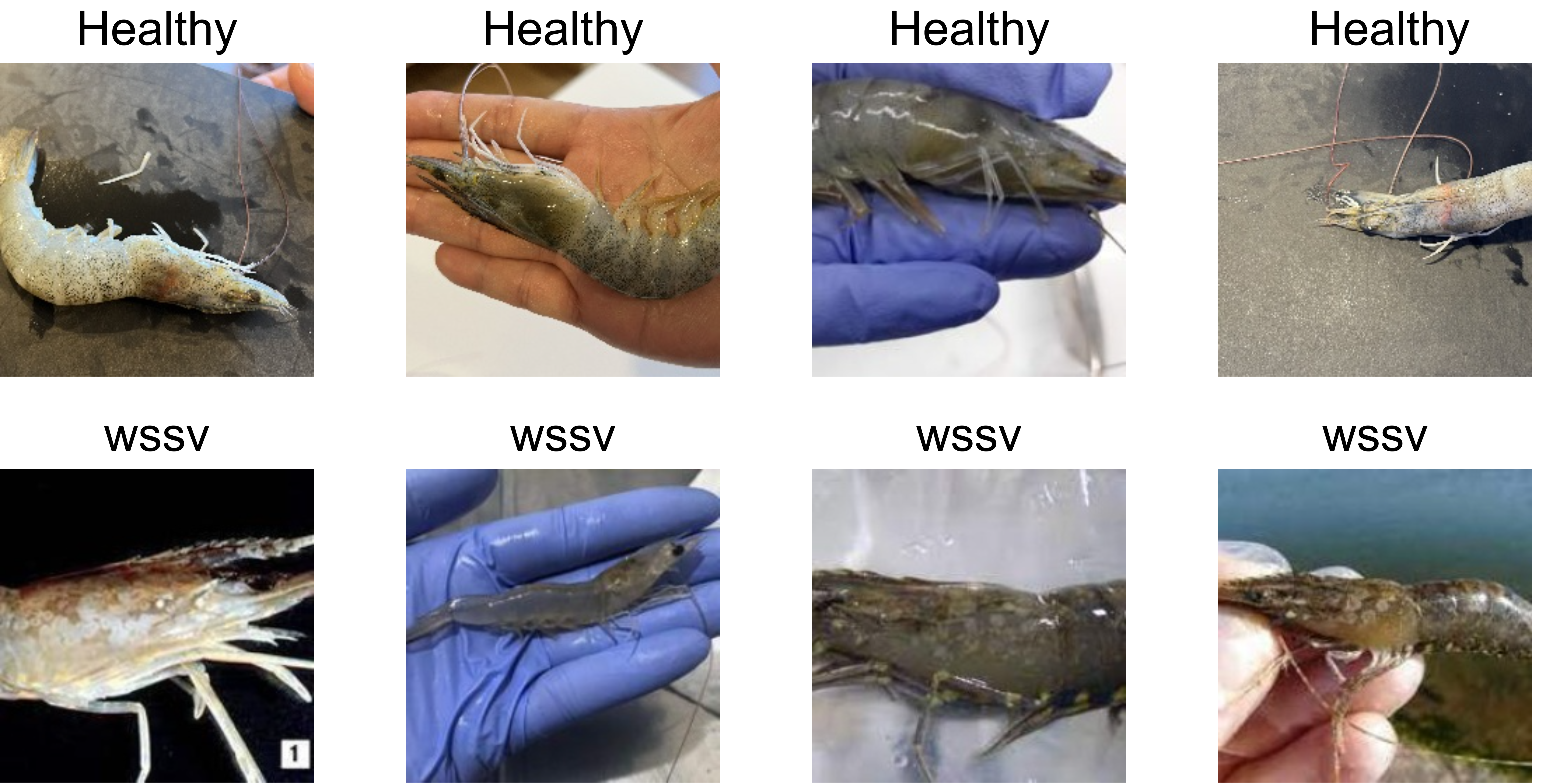}
    \caption{Image samples from the dataset. The top row contains healthy images, while the bottom row contains WSSV-infected images.}
    \label{fig:dataset-samples}
\end{figure}

The performance of two model architectures, MobileNetV3-Small and EfficientNetV2-B0, were analyzed for their feasibility for on-device inference. The model architectures are recognized for their balance in performance and compact model size, making them well-suited for deployment on resource-constrained devices like smartphones and microcontrollers \cite{facial2021, lowpower2020}. Both models leverage transfer learning. Fine-tuning was performed by unfreezing the last specialized layers that were initially pretrained on the ImageNet dataset. This process allowed the models to adapt their performance for the WSSV recognition task. The models were trained for a total of 300 epochs and a batch size of 64 per iteration, with an early stopping callback implemented to prevent overfitting. The learning rate was set to 1e-6 while using Adam as the optimizer, with Binary Cross-Entropy as the loss function.

Evaluation metrics include F$_1$-Score, Area Under the Curve - Receiver Operating Characteristic (AUC-ROC), and False Negative Rate (FNR). These metrics were selected as they are known to thoroughly describe the performance of models in the context of imbalanced classification \cite{he2013imbalanced}.

\section{Experiments}
The experiments in this study were designed to provide a thorough analysis of the selected models, allowing for both quantitative and qualitative evaluation. Computer vision-related experiments were conducted on a Windows 11 64-bit system equipped with an AMD Ryzen 2700X processor, 32GB of RAM, and an Nvidia Geforce RTX 3060 GPU. The deep learning models were trained on the TensorFlow framework v2.12.0.

\subsection{Quantitative Analysis}
In this study, stratified 5-fold cross-validation was used. Cross-validation addresses the issue of the limited dataset size, allowing for a more reliable evaluation of the model's performance. Table \ref{tab:metrics} presents the evaluation results obtained from the 5-fold cross-validation. 

The EfficientNetV2-B0 model consistently outperformed the MobileNetV3-Small model across all splits. In the training split, the EfficientNetV2-B0 model achieved an F$_1$-Score of 0.99, indicating its ability to effectively balance precision and recall. Additionally, it attained an AUC value of 1.00, indicating ideal discriminative power. The EfficientNetV2-B0 model also gained an FNR of 0.00. In the validation split, the Efficient\-NetV2-B0 model maintained its remarkable performance with an F$_1$-Score of 0.93, an AUC value of 1.00, and achieved an FNR of 0.00, also indicating a good hit rate in identifying WSSV-infected shrimp.

\begin{table}[!t]
\renewcommand{\arraystretch}{1.75}
\centering
\caption{Evaluation Results from the 5-fold cross-validation}
\label{tab:metrics}
\begin{tabular}{llcc}
\hline
\textbf{Split} & \textbf{Metrics}                      & \multicolumn{2}{c}{\textbf{Model}}    \\ \hline 
               &                                       & MobileNetV3-Small & EfficientNetV2-B0 \\ \cline{3-4} 
Training       & F$_1$-Score $\uparrow$     & 0.72 $\pm$ 0.25           & \textbf{0.99 $\pm$ 0.01}         \\
               & AUC $\uparrow$             & 0.99 $\pm$ 0.00           & \textbf{1.00 $\pm$ 0.00}         \\
               & FNR $\downarrow$           & 0.09 $\pm$ 0.15           & \textbf{0.00 $\pm$ 0.00}         \\ \hline
Validation     & F$_1$-Score $\uparrow$     & 0.58 $\pm$ 0.33           & \textbf{0.93 $\pm$ 0.09}         \\
               & AUC $\uparrow$             & 0.99 $\pm$ 0.00           & \textbf{1.00 $\pm$ 0.00}         \\
               & FNR $\downarrow$           & 0.36 $\pm$ 0.31           & \textbf{0.00 $\pm$ 0.00}         \\ \hline
\end{tabular}
\end{table}

\begin{figure}[!t]
    \centering
    \includegraphics[width=3.45in]{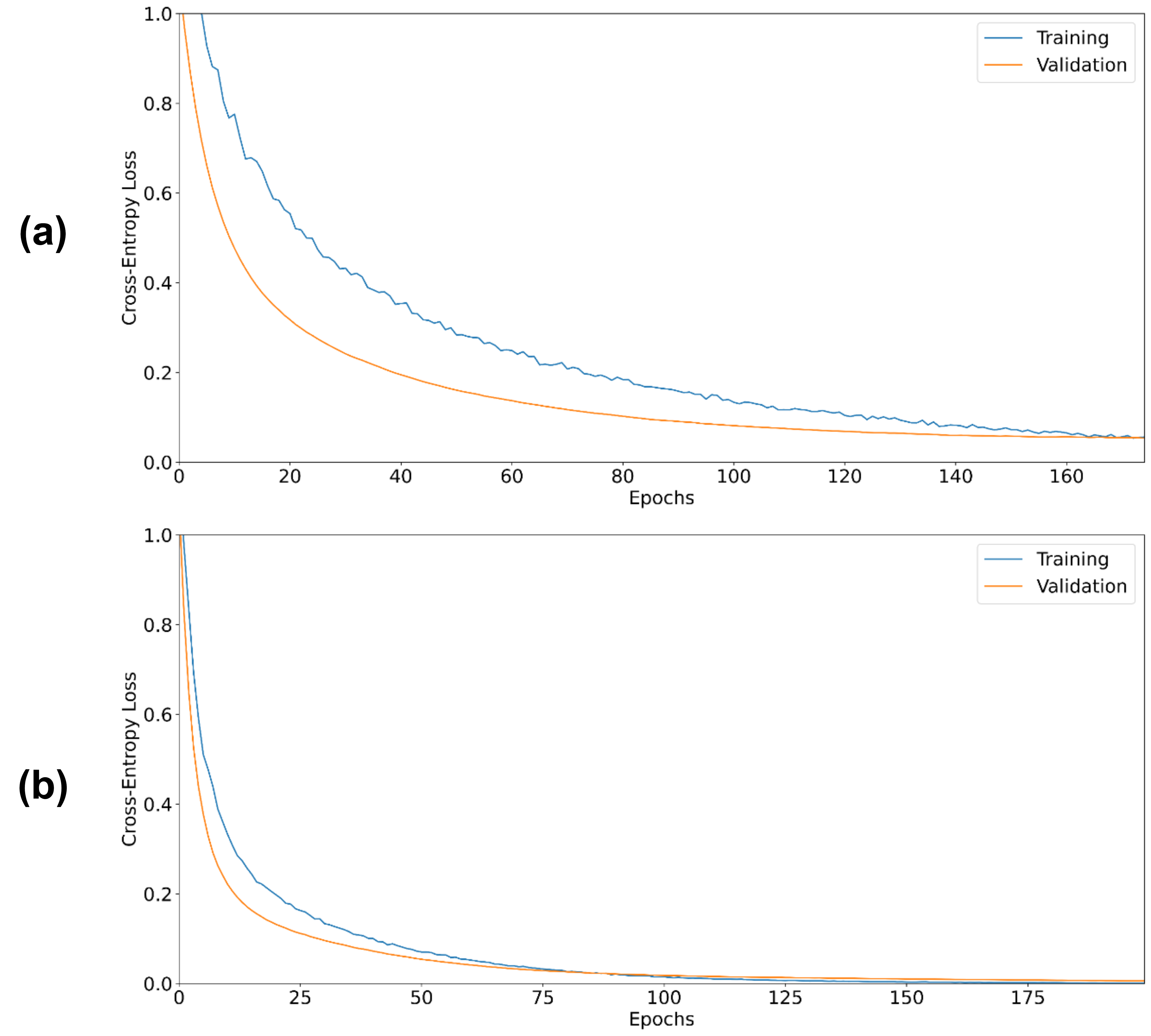}
    \caption{The training and validation loss curves of the (a) MobileNetV3-Small model, and (b) EfficientNetV2-B0 model.}
\end{figure}

Overall, the EfficientNetV2-B0 model demonstrated remarkable performance in terms of F$_1$-Score and AUC across all validation splits, indicating its effectiveness in WSSV recognition. It is also important to consider that the performance on the test set may be influenced by the limited dataset size and class imbalance, magnifying the weight of each misclassification. Nonetheless, the evaluation metrics provide valuable insights into the models' performance and their potential for mobile application deployment.

\begin{figure}[!t]
    \centering
    \includegraphics[width=3.45in]{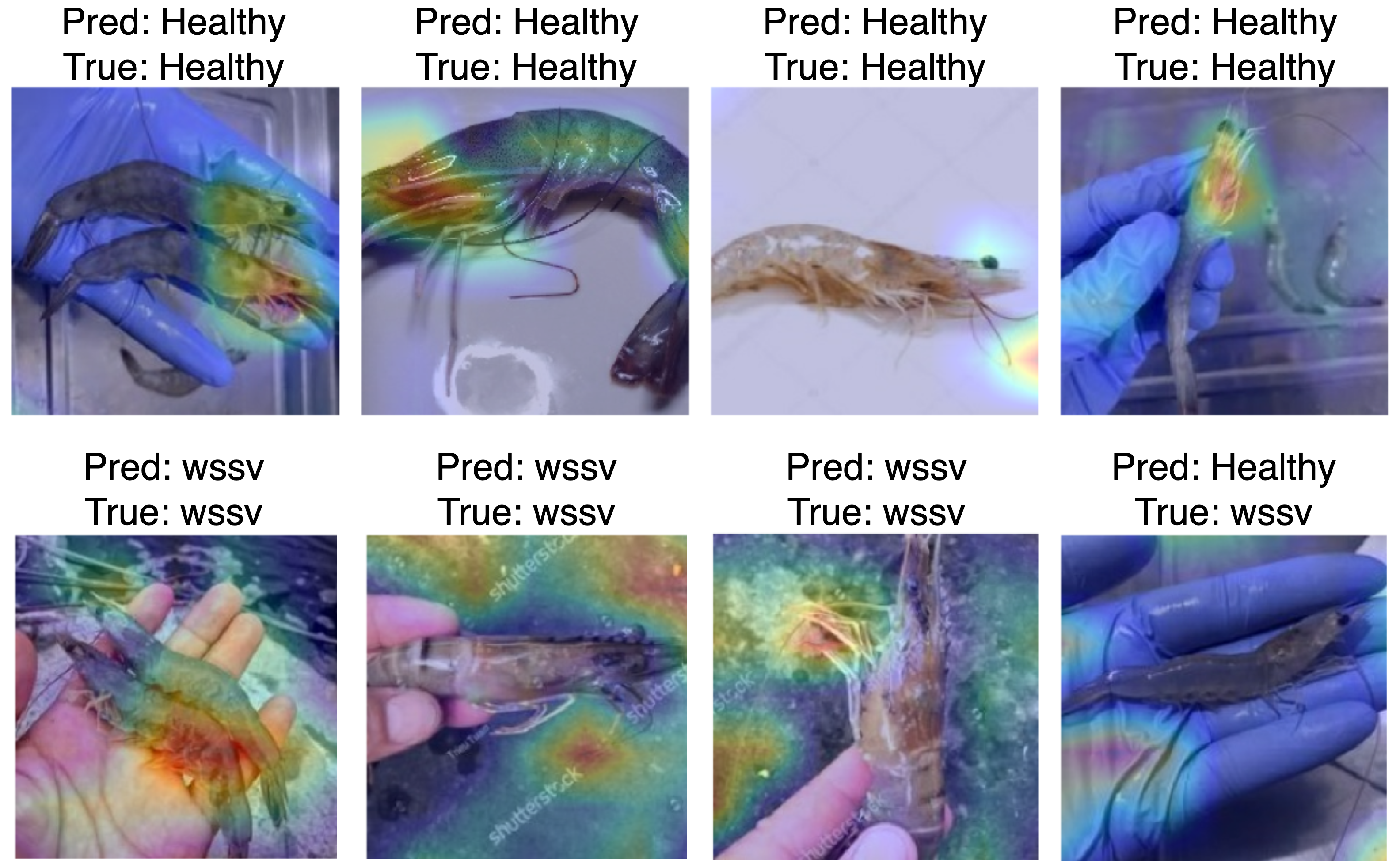}
    \caption{Saliency heatmaps generated by the MobileNetV3-Small model. The top row is image samples of healthy shrimps, while the bottom row is WSSV-infected shrimps. It can be observed that the model struggles to focus on the region of interest, indicating the model not effectively capturing the distinguishing features associated with WSSV-infected shrimps and would need further investigation on its lack of attention.}
    \label{fig:mobilenet-saliency}
\end{figure}

\begin{figure}[!t]
\centering
\includegraphics[width=2in]{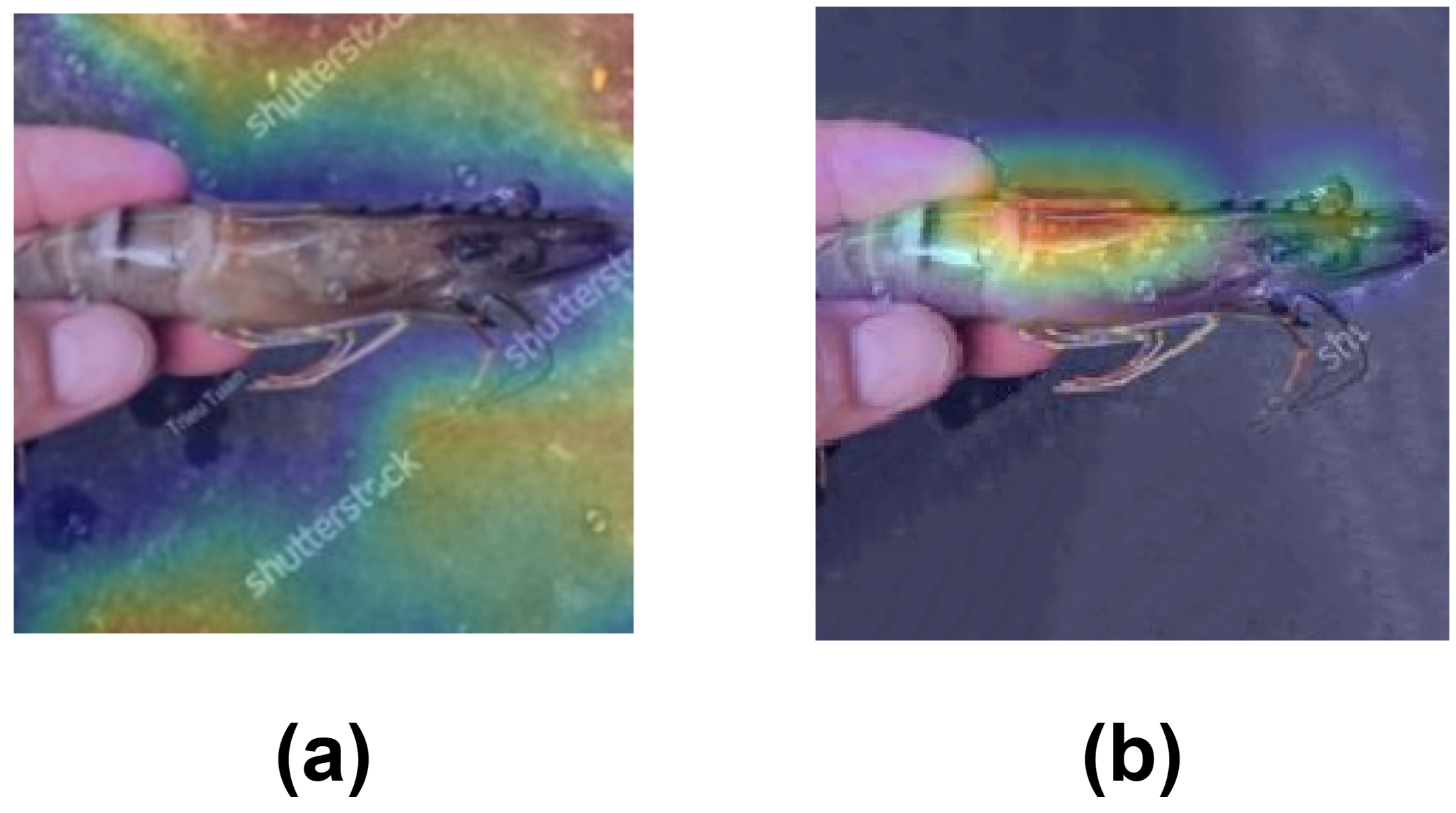}
\caption{By removing noise that resembled WSSV symptoms, the shift in attention of the MobileNetV3-Small model can be clearly observed. The model's focus is primarily on the shrimp itself, indicating that the background was a major factor in diverting the model's attention away from the region of interest. This finding suggests that the model may have been confused by the presence of speckles in the background.}
\label{fig:mobilenet-bg-effect}
\end{figure}

\subsection{Qualitative Analysis}
The general observation from the analysis of the MobileNetV3-Small and EfficientNetV2-B0 model's behavior, as seen in Fig. \ref{fig:efficientnet-saliency} and Fig. \ref{fig:mobilenet-saliency}, is that EfficientNetV2-B0 model performs much better in capturing and focusing on the shrimp images, indicating that the model is learning much better in the meaningful patterns found in each image. In most cases, the saliency heatmaps generated by the MobileNetV3-Small model (Fig. \ref{fig:mobilenet-saliency}) suggest that the model does not prioritize focusing on the shrimp image samples, especially those exhibiting symptoms of WSSV. This lack of attention to the shrimp in the heatmaps implies that the model may be relying on other visual cues or patterns in the images that are not directly related to the shrimp itself. This could explain the MobileNetV3-Small model generally performing worse based on the quantitative analysis (see Table \ref{tab:metrics}). Other possible reasons for the MobileNetV3-Small model's lack of focus on the shrimp in the saliency heatmaps could be the limited diversity of WSSV-labeled images in the training dataset.

Additionally, the presence of different backgrounds could contribute to the generalizability of the model. As seen in Fig. \ref{fig:mobilenet-bg-effect}a, a background that resembles the pattern of WSSV symptoms caused the model to ignore the shrimp in the foreground. However, once the background was edited to remove the white speckles (as illustrated in Fig. \ref{fig:mobilenet-bg-effect}b), the model shifted its focus. Including segmentation during training could aid in helping the model learn better from the region of interest, but traditional segmentation techniques may struggle to accurately separate the shrimp from complex backgrounds, leading to confusion in the model's attention. Exploring and incorporating deep learning-based segmentation techniques into the training pipeline could help improve the model's ability to distinguish shrimp from the background.

\subsection{Model Conversion Analysis}
Observing the discrepancies that occur in the conversion process from TensorFlow to ONNX is crucial to ensure the accuracy and reliability of the converted model. One possible discrepancy is related to the supported operations and their implementations in the target framework. Another discrepancy that can occur is in the numerical precision of the model. These differences may be negligible in some cases, but in certain scenarios, such as when dealing with highly sensitive or critical applications, they can have significant implications.

\begin{table}[!t]
\renewcommand{\arraystretch}{1.75}
\centering
\caption{Absolute error of Sigmoid activation output between the models}
\label{tab:model-error}
\begin{tabular}{@{}lcccc@{}}
\hline
\textbf{Model}    & \multicolumn{4}{c}{\textbf{Absolute Error}} \\ \hline
                  & Mean        & Stddev      & Min     & Max        \\ \cline{2-5} 
MobileNetV3-Small & 8.88e-06    & 3.80e-05    & 0.00    & 2.68e-04   \\
EfficientNetV2-B0 & 3.63e-05    & 2.09e-04    & 0.00    & 1.65e-03   \\ \hline
\end{tabular}
\end{table}

Table \ref{tab:model-error} displays the absolute error of the Sigmoid activation output between the original TensorFlow models and their corresponding ONNX-converted models. The absolute error is calculated as the absolute value of the difference between the Sigmoid activation outputs of the models. The metrics provided include the mean, standard deviation, minimum, and maximum values. The MobileNetV3-Small model exhibited a mean absolute error of 8.88e-06, with a standard deviation of 3.80e-05. The EfficientNetV2-B0 model had a slightly higher mean absolute error of 3.63e-05, with a standard deviation of 2.09e-04. The observed discrepancies between the TensorFlow and ONNX models can be linked to the quantitative evaluation results, particularly the AUC-ROC metric. Minimal discrepancies in the model output during the conversion process suggest that the performance of the ONNX models is likely to be comparable to that of the original TensorFlow models. This is supported by the high AUC values obtained during the evaluation, indicating the models' strong discriminative power and ability to differentiate between healthy and WSSV-infected shrimp samples.

\subsection{On-Device Inference Results}
To assess the models' performance in a real-world deployment scenario, on-device inference evaluations were conducted. These evaluations involved running the trained models in inference mode 5 times and recording the average latency of the predictions. Note that these evaluations were performed on the CPU to account for devices that may not have a dedicated GPU for computationally intensive tasks.

\begin{table}[!t]
\renewcommand{\arraystretch}{1.75}
    \centering
    \caption{Inference times of both trained models in each smartphone.}
    \label{tab:inference_time}
    \begin{tabular}{@{}lcc@{}}
    \hline
    \textbf{Smartphone Model}     & \multicolumn{2}{c}{\textbf{Inference Time (CPU)}} \\ \hline
                                  & MobileNetV3-Small       & EfficientNetV2-B0       \\ \cline{2-3} 
    Xiaomi Mi 11 Lite (2021)      & \textbf{406.98 ms}      & \textbf{422.12 ms}      \\
    Galaxy Note 10 (2020) & 412.56 ms               & 456.64 ms               \\
    Samsung S10 (2019)            & 423.43 ms               & 478.12 ms               \\ \hline
    \end{tabular}
\end{table}

The results shown in Table \ref{tab:inference_time} revealed that the MobileNetV3-Small model generally exhibited lower inference latency compared to the EfficientNetV2-B0 model, indicating faster prediction times by a few milliseconds. Furthermore, a general trend was observed in which the latency of the models appeared to increase for smartphones purchased earlier than the testing period. This finding suggests that older smartphone models with less advanced hardware specifications may have slightly higher inference latency than newer models, emphasizing the consideration of the device capabilities and hardware specifications when deploying models on low-power devices.

\section{Conclusion}
Limited progress has been made in establishing a monitoring and early reporting system in the aquaculture industry, especially utilizing widely available technology like smartphones. This study presented a mobile application that facilitates data collection and monitoring efforts to address the absence of an open-source dataset. This study collected images for training the WSSV recognition model. As the number of collected images increases, the dataset could potentially serve as a valuable resource for future research and development in this domain.

To showcase the potential of the collected images in WSSV detection, a computer vision-based model was developed and evaluated. Data augmentation schemes that simulate different lighting conditions and camera orientation are used to increase the number of training samples. To better reflect the performance of the model trained and tested on the imbalanced dataset, appropriate performance metrics were monitored.

The developed system will benefit the aquaculture industry and regulatory bodies, facilitating informed decisions, disease prevention, and sustainable practices. The collected data could support further studies in shrimp disease management and contribute to effective strategies for WSSV prevention and control. Future directions may involve monitoring WSSV challenge experiments, capturing images at regular intervals to identify early signs of WSSV, and exploring intervention efficacy. Such experiments provide valuable insights, enhancing the accuracy and reliability of the WSSV recognition model for more precise and timely disease detection.

\section*{Acknowledgment}
The authors would like to thank the Bureau of Fisheries and Aquatic Resources (BFAR) and the shrimp farms that provided assistance in collecting shrimp image samples.

\bibliographystyle{IEEEtran}
\bibliography{cites}

\end{document}